\pdfoutput=1

\documentclass[11pt]{article}

\usepackage[final]{acl}

\usepackage{times}
\usepackage{latexsym}

\usepackage[T1]{fontenc}

\usepackage[utf8]{inputenc}

\usepackage{microtype}

\usepackage{inconsolata}

\usepackage{graphicx}
\usepackage{microtype}
\usepackage{graphicx}
\usepackage{subfigure}
\usepackage{booktabs} 
\usepackage{multirow}
\usepackage{manyfoot}
\usepackage{tabularx}
\usepackage{amsmath}
\usepackage{amssymb}
\usepackage{mathtools}
\usepackage{amsthm}
\usepackage{algorithm}
\usepackage{algpseudocode}
\usepackage{amsfonts} 
\title{Riemannian Optimization for LoRA on the Stiefel Manifold}

\author{
Juneyoung Park$^{*}$, Minjae Kang$^{*}$, Seongbae Lee$^{*}$, \\ \textbf{Haegang Lee}, \textbf{Seongwan kim}\textsuperscript{$\dagger$}, \textbf{Jaeho Lee}\textsuperscript{$\dagger$}
\\ \\ Opt-AI Inc. \\
}

\def \name{Stiefel-LoRA}
\begin{document}
\maketitle
\renewcommand{\thefootnote}{\fnsymbol{footnote}}
\setcounter{footnote}{1}
\footnotetext{Equal Contribute}
\renewcommand{\thefootnote}{\fnsymbol{footnote}}
\setcounter{footnote}{2}
\footnotetext{Corresponding Author}
\setcounter{footnote}{3}
\footnotetext{\texttt{\{jyoung.park, mjae.kang, sbae.lee, hgang.lee, swan.kim, jaeho.lee\}@opt-ai.kr}}
\begin{abstract}
While powerful, large language models (LLMs) present significant fine-tuning challenges due to their size. Parameter-efficient fine-tuning (PEFT) methods like LoRA provide solutions, yet suffer from critical optimizer inefficiencies; notably basis redundancy in LoRA's $B$ matrix when using AdamW, which fundamentally limits performance. We address this by optimizing the $B$ matrix on the Stiefel manifold, imposing explicit orthogonality constraints that achieve near-perfect orthogonality and full effective rank. This geometric approach dramatically enhances parameter efficiency and representational capacity. Our Stiefel optimizer consistently outperforms AdamW across benchmarks with both LoRA and DoRA, demonstrating that geometric constraints are the key to unlocking LoRA's full potential for effective LLM fine-tuning.
\end{abstract}

\section{Introduction}
\begin{figure*}[h]
\begin{center}
\centerline{\includegraphics[width=\textwidth]{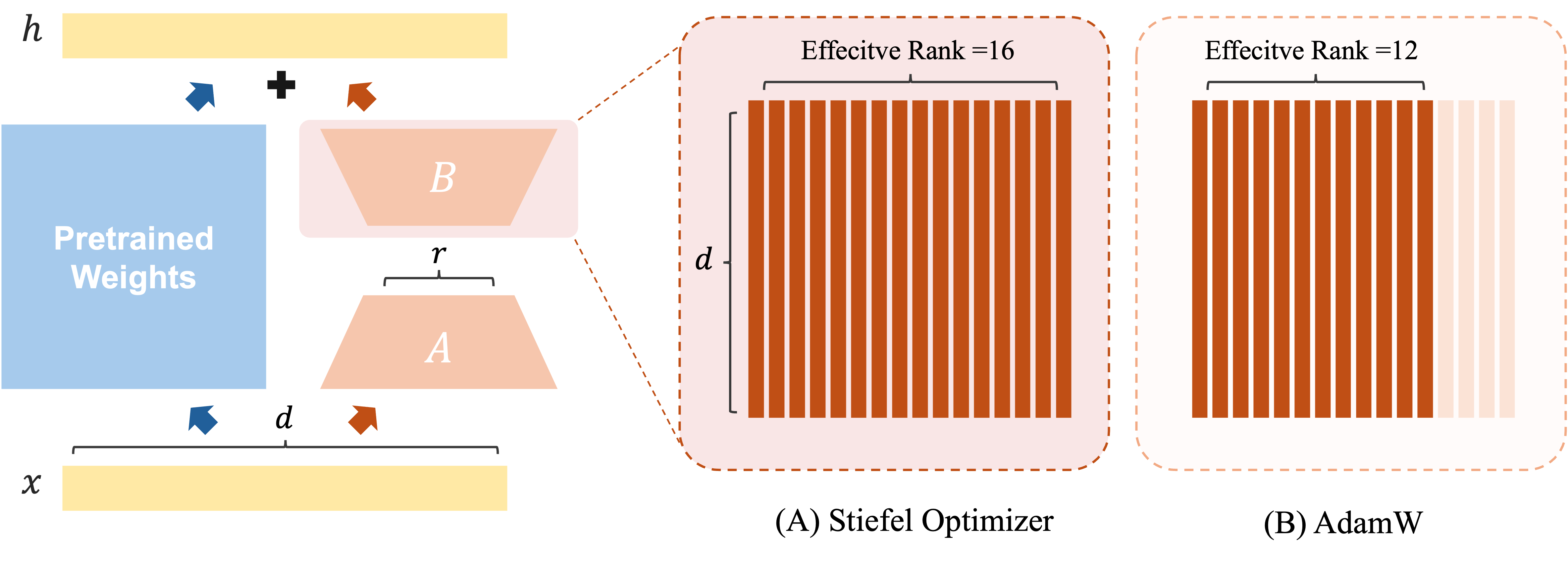}}
\caption{Visualization of the effective rank of LoRA weights with and without the Stiefel manifold constraint, illustrated for the case of r=16. 
(A) When the LoRA matrix $B \in \mathbb{R}^{d \times r}$ is optimized on the Stiefel manifold, its columns remain orthogonal. This ensures linear independence and allows them to fully span an r-dimensional subspace, resulting in an effective rank equal to the nominal rank r. 
(B) Without this orthogonal structure (e.g., when trained in Euclidean space with AdamW), the columns of B can become correlated or redundant, leading to an effective rank lower than the nominal rank r.}
\label{fig1}
\end{center}
\vskip -0.2in 
\end{figure*}
Large Language Models (LLMs) have recently led to significant progress in the field of Natural Language Processing (NLP), achieving near or superhuman performance across diverse tasks \cite{brown2020language, touvron2023llama2openfoundation}. However, the substantial computational overhead and memory footprint associated with LLMs, which possess parameters numbering hundreds of billions, impose significant constraints on their widespread deployment and efficient fine-tuning \cite{kaplan2020scaling}. To mitigate these pragmatic limitations and effectively adapt LLMs for various downstream tasks, Parameter-efficient fine-tuning (PEFT) techniques have emerged, which involve fine-tuning only a minimal subset of the original model weights \cite{lialin2023scaling, lester2021powerscaleparameterefficientprompt, li2021prefixtuningoptimizingcontinuousprompts}. Notably, Low-Rank Adaptation (LoRA) \cite{hu2022lora} represents the weight update $\nabla W$ as the product of two low-rank matrices, $A\in \mathbb{R}^{r \times k}$ and $B \in \mathbb{R}^{d \times r}$ (i.e.,$\nabla W = BA$), thereby drastically reducing the total number of trainable parameters. This approach efficiently achieves performance analogous or competitive to full model fine-tuning, establishing LoRA as one of the most ubiquitously adopted PEFT methodologies at present.

Notwithstanding the success of LoRA, a fundamental question arises: \textit{Are we truly leveraging this constrained resource $r$, in the most effective way?} Prevailing LoRA implementations predominantly train the update matrices $A$ and $B$ within a standard Euclidean space using conventional gradient descent-based optimization algorithms, for example, AdamW, without imposing explicit structural constraints \cite{loshchilov2017decoupled}. While this approach offers straightforward implementation, it may overlook potential inefficiencies stemming from the inherent low-dimensional structure. For instance, during the training process, the column vectors of matrix $B$, which are the basis directions of the update, may exhibit increased similarity, leading to redundancy. Alternatively, the update directions might manifest unstable dynamics. Such phenomena can result in an underutilization of the representational capacity afforded by the fixed rank $r$. This, in turn, can cause slower convergence, suboptimal final performance, or require a higher rank $r$, and consequently more parameters, to achieve satisfactory performance levels \cite{kalajdzievski2023rankstabilizationscalingfactor}.

In stark contrast, within other domains of deep learning research, the imposition of appropriate geometric structures, particularly orthogonality or unitarity constraints, on learnable parameter matrices has been consistently reported to be highly effective in enhancing model performance and training stability. For example, in Convolutional Neural Networks (CNNs), enforcing orthogonality constraints on weight matrices has been demonstrated to bolster the orthogonality of feature representations, thereby enriching their expressiveness and consequently improving the model's generalization capabilities and training stability \cite{bansal2018can, huang2018orthogonal, wang2020orthogonal}. In Recurrent Neural Networks (RNNs), the application of unitary matrix constraints, which are orthogonal matrices in complex space, has proven effective in mitigating the vanishing and exploding gradient problems encountered during the learning of long-term dependencies \cite{arjovsky2016unitary, vorontsov2017orthogonality}. These diverse success stories strongly suggest that introducing suitable geometric structural constraints into the model parameter space, aligned with the characteristics of the data or the learning objectives, can exert a positive influence on the model's representational power, learning dynamics, and ultimate performance.

This then leads to a natural inquiry: \textit{Could such geometric constraints be effectively applied to the low-rank update matrices in LoRA, the cornerstone of PEFT techniques, to overcome the aforementioned limitations and unlock latent performance capabilities?} This research endeavors to address this pivotal question. We propose \name, a novel fine-tuning framework that explicitly imposes orthogonality constraints on the update matrix $B$, a core component of LoRA. Specifically, we constrain the column vectors of matrix $B$ to be orthonormal, which is tantamount to $B$ residing on the Stiefel manifold $St(d,r)={B\in \mathbb{R}^{d\times r}:B^{\top}B=I_{r}}$ \cite{absil2009optimization, edelman1998geometry}. Under this constraint, the LoRA update $\nabla W = BA$ can be interpreted as a linear transformation within the space spanned by the orthonormal column vectors of matrix $B$. We anticipate that the orthogonality constraint imposed in Stiefel-LoRA will maximize the representational efficiency of the LoRA update, eliminate unnecessary redundancies, and stabilize learning dynamics, thereby eliciting the maximum potential performance of LoRA under a given parameter budget (rank $r$). This will ultimately lead to achieving comparable or superior performance with fewer parameters ($r$), or attaining faster convergence and higher final performance with the same number of parameters. 

This paper delineates the theoretical underpinnings of the Stiefel-LoRA framework and provides an in-depth analysis of how the proposed geometric constraint contributes to ameliorating the inherent limitations of standard LoRA, particularly concerning parameter efficiency and training stability. Furthermore, through extensive experimentation across diverse LLMs fine-tuning benchmarks, we successfully demonstrate that Stiefel-LoRA consistently achieves superior performance, markedly faster convergence rates, and enhanced parameter efficiency compared to standard LoRA. In conclusion, this research introduces a novel perspective of geometric optimization to parameter-efficient fine-tuning techniques, thereby paving new avenues for optimizing the efficiency and performance of LLMs fine-tuning and is anticipated to contribute to the advancement of related research.


\section{Related Works}
\subsection{PEFT and LoRA}
Fine-tuning entire large-scale models demands substantial computational resources and memory. To address these challenges, PEFT methodologies aim to achieve performance comparable to full fine-tuning by optimizing only a small number of parameters while keeping the majority of the pre-trained model weights frozen \cite{lester2021powerscaleparameterefficientprompt, li2021prefixtuningoptimizingcontinuousprompts}. LoRA stands as a prominent PEFT methodology, which fine-tunes by adding low-rank adapters to the weight matrices of a pre-trained model.

Research on LoRA has concentrated on enhancing training stability, conducting comparative analyses with full fine-tuning, and developing various derivative models to address the issue of catastrophic forgetting. For instance, rsLoRA \cite{kalajdzievski2023rankstabilizationscalingfactor} (Rank-Stabilized LoRA) modifies the scaling factor of conventional LoRA to resolve the gradient collapse problem at high ranks, thereby improving training stability. This allows the model to effectively leverage higher ranks to enhance performance.

Comparative studies between LoRA and Full fine-tuning (FFT) have elucidated structural and behavioral distinctions between the two approaches. LoRA generally exhibits a reduced tendency to forget pre-learned knowledge compared to FFT. \cite{biderman2024lora} However, a performance gap may be observed in certain complex domains \cite{tian2024hydralora, liuseedlora}. Furthermore, a novel phenomenon termed "intruder dimensions" has been observed in LoRA-tuned models, which are not present in FFT and may affect robustness during continual learning. \cite{reece2024lora}

To overcome these limitations and mitigate catastrophic forgetting, variant models such as DoRA \cite{liu2024doraweightdecomposedlowrankadaptation} (Weight-Decomposed Low-Rank Adaptation), LoRA-Null \cite{tang2025lora}, and LoRAX \cite{sullivan2025lorax} (LoRA eXpandable Networks) have been proposed. DoRA decomposes pre-trained weights into magnitude and direction, then applies LoRA to update the directional component, aiming to emulate the learning capacity of full fine-tuning. LoRA-Null aims for knowledge preservation by initializing LoRA adapters in the null space of pre-trained knowledge activation. LoRAX balances stability and plasticity in continual learning environments by adding new LoRA modules for each task.

Close to our work, \citet{buyukakyuz2024olora} optimizes learning through orthogonality via QR Decomposition. However, the primary goal of that study is convergence speed, rather than performance optimization. Furthermore, our research differs in that it utilizes Riemannian optimization, leveraging the advantages of the Stiefel manifold. 

\subsection{Riemannian Geometric Optimization}
In deep learning model training, conventional optimization methodologies predicated on Euclidean space exhibit limitations in fully harnessing the complex geometric structures inherent in parameter spaces \cite{martens2020new, fei2025survey}. As an alternative, Riemannian geometry-based optimization methodologies are garnering attention, as they can leverage the intrinsic geometric information of data or model parameters to enhance learning dynamics and deepen theoretical understanding \cite{zhang2016first, absil2009optimization}.

Riemannian Stochastic Gradient Descent (RSGD) extends Stochastic Gradient Descent (SGD) to Riemannian manifolds \cite{bonnabel2013stochastic}. Research indicates that RSGD can achieve faster convergence rates when employing increasing batch sizes alongside gradually decreasing learning rates, a trend analogous to findings in Euclidean SGD \cite{goyal2017accurate, smith2017don}. Riemannian optimization can facilitate convergence by transforming constrained optimization problems into unconstrained problems on a manifold, sometimes attaining higher accuracy than Euclidean methods \cite{huang2017riemannian}. Riemannian Bilevel Optimization (RieBO, RieSBO) algorithms have been shown to achieve gradient complexity and oracle call counts similar to their Euclidean counterparts \cite{li2025riemannian}.


Although LoRA and Riemannian optimization originate from different perspectives, they share a common objective of achieving efficient learning in high-dimensional, complex spaces. LoRA contributes to this goal by streamlining the learning process itself \cite{hu2022lora}, while Riemannian optimization does so by enabling a more efficient exploration of the parameter space through the utilization of its geometric properties \cite{absil2009optimization}. Currently, the optimization of LoRA parameters is predominantly performed in Euclidean space. However, if LoRA parameters themselves reside on a manifold satisfying specific constraints, or if their update process could benefit from geometric considerations, applying Riemannian optimization techniques to LoRA parameter optimization could represent a promising research direction.
\section{Methodology}
This chapter explains the core concepts of the proposed Stiefel-LoRA methodology. The basic concepts of LoRA, the Stiefel manifold, and Riemannian optimization are described in Appendix \ref{app:preliminaries}. Then, Algorithm \ref{alg:stiefel_lora_qr_adam} presents the core mechanism and the overall algorithm for efficiently applying the Stiefel manifold constraint to LoRA updates.

\subsection{Proposed Method}
\begin{table*}[ht]
\setlength{\tabcolsep}{1.2mm}
\centering
\resizebox{0.99\textwidth}{!}{
\begin{tabular}{cccccccccccccc}
\toprule
\textbf{Model} & \textbf{Method} & \textbf{Optimizer} & \textbf{BoolQ} & \textbf{PIQA} & \textbf{SIQA} & \textbf{HellaSwag} & \textbf{ARC-e} & \textbf{ARC-c} & \textbf{OBQA} & \textbf{Avg.}\\
\toprule
\midrule

\multirow{4}{*}{LLaMA3.2-1B} 
& \multirow{2}{*}{LoRA} & \textbf{Stiefel} & 75.2 & 70.9 & 65.3 & 29.2 & 70.5 & 44.2 & 63.2 & \textbf{59.7} \\
&                        & AdamW  & 63.2 & 53.4 & 50.1 & 25.4 & 58.8 & 35.7 & 46.6 & 47.6 \\ 
& \multirow{2}{*}{DoRA} & \textbf{Stiefel} & 77.5 & 71.4 & 66.8 & 30.5 & 71.2 & 37.6 & 64.7 & \textbf{59.9} \\
&                        & AdamW  & 67.6 & 65.2 & 61.4 & 26.9 & 60.5 & 36.4 & 48.1 & 52.3 \\

\midrule

\multirow{4}{*}{LLaMA3.2-3B} 
& \multirow{2}{*}{LoRA} & \textbf{Stiefel} & 84.7 & 85.1 & 82.5 & 90.3 & 85.4 & 68.6 & 80.4 & \textbf{82.4} \\
&                        & AdamW  & 81.1 & 80.5 & 78.9 & 87.1 & 83.4 & 65.1 & 76.2 & 78.9 \\
& \multirow{2}{*}{DoRA} & \textbf{Stiefel} & 86.5 & 87.1 & 84.4 & 92.5 & 87.9 & 70.8 & 82.7 & \textbf{84.5} \\
&                        & AdamW  & 83.8 & 82.2 & 80.6 & 89.7 & 85.2 & 67.7 & 78.5 & 81.1 \\

\midrule

\multirow{4}{*}{LLaMA3-8B} 
& \multirow{2}{*}{LoRA} & \textbf{Stiefel} & 86.2 & 87.9 &  82.8  & 91.5  & 87.2  & 72.1  & 81.9 & \textbf{84.2}  \\
&                        & AdamW  & 83.3 & 81.5 &  78.8  & 88.3  & 85.4 & 68.2 & 77.7  & 80.4 \\
& \multirow{2}{*}{DoRA} & \textbf{Stiefel} & 88.7   & 89.5  & 85.1   & 94.2 & 89.4  & 74.6   & 84.9  & \textbf{86.6}  \\
&                        & AdamW  & 85.9  & 83.8  & 81.1 & 90.9  & 87.2  & 71.2  & 80.3  & 82.9 \\

\bottomrule
\end{tabular}
}
\caption{Accuracy comparison on seven commonsense reasoning datasets with various PEFT($r=16$) method and optimizer applied.}
\label{tab:llama_commonsense}
\end{table*}

Stiefel-LoRA aims to enhance the LoRA fine-tuning methodology by explicitly imposing an orthogonality constraint on one of the factor matrices composing the low-rank approximation of a weight matrix, thereby augmenting parameter efficiency and improving model performance. The core of this approach lies in performing optimization on the Stiefel manifold, which helps to overcome potential limitations in the expressive power of conventional LoRA methods and fosters a more stable learning process.

The optimization problem for Stiefel-LoRA is formulated as follows for a given fine-tuning loss function $f$:
\begin{equation}
\min_{A,B} f(W_{0} + BA)\ \text{subject to}\ B\in St(n,p) \label{eq:stiefel_lora_problem}
\end{equation}
Here, $W_0 \in \mathbb{R}^{d \times k}$ represents the pre-trained weight matrix, which remains fixed throughout the fine-tuning process. The matrix $A \in \mathbb{R}^{r \times k}$, one of the optimization targets, is searched within the standard Euclidean space without any additional constraints. Conversely, the other target matrix, $B \in \mathbb{R}^{d \times r}$, is optimized on the Stiefel manifold $St(d,r)$. In this context, the constraint $B \in St(d,r)$ (implying $B^{\top}B = I_{r}$, where $I_{r}$ is the $r \times r$ identity matrix) enforces that the column vectors of $B$ are orthonormal.

The initial step in the optimization process involves computing the Euclidean gradients of the loss function $f$ with respect to $A$ and $B$, denoted as $\nabla_A f$ and $\nabla_B f$, respectively, using the standard backpropagation algorithm. The gradient $\nabla_A f$ for matrix $A$, which is optimized in Euclidean space, is directly used to update $A$ in the conventional manner. However, for matrix $B$, which must satisfy the Stiefel manifold constraint, the Euclidean gradient $\nabla_B f$ is not directly used for updates. Instead, it (or a momentum-updated version thereof) undergoes a projection onto the tangent space at the current point $B_{k}$ on the manifold to form a tangent vector $\xi_k$. This tangent vector $\xi_k$, scaled by a learning rate $\alpha$, is then utilized in a retraction operation to move from $B_{k}$ to the next point $B_{k+1}$ that satisfies the manifold constraint.

For the retraction operation, which maps a point $B_{k} \in St(d,r)$ and a tangent vector $\xi_{k} \in \mathcal{T}_{B_{k}} St(d,r)$ to a new point $B_{k+1} \in St(d,r)$, we employ a method based on QR decomposition. This common retraction, often referred to as projection via QR decomposition, is performed in two steps;
\begin{itemize}
    \item[1.] An `optimistic' step is taken in the ambient Euclidean space $\mathbb{R}^{d \times r}$  along the tangent direction: $\mathcal{Y}^{'}_{k}=B_{k}+\alpha \xi_{k}$
    \item[2.] The resulting matrix $\mathcal{Y}^{'}_{k}$ generally does not lie on the Stiefel manifold (i.e., its columns may not be orthonormal). It is projected back to $St(d,r)$ by performing its QR decomposition. If $\mathcal{Y}^{'}_{k}=Q_{k}R_{k}$ is the QR decomposition of $\mathcal{Y}^{'}_{k}$(where $Q_{k} \in \mathbb{R}^{d \times r}$ has orthonormal columns and $R_{k} \in \mathbb{R}^{r \times r}$ is upper triangular), the new point is taken as $B_{k+1}=Q_{k}$
\end{itemize}
To ensure uniqueness and desirable properties for $Q_{k}$(such as forming a valid retraction), variants of QR decomposition can be used where the diagonal elements of $R_{k}$ are constrained to be positive. This QR-based retraction robustly ensures that $B_{k+1}$ satisfies the orthonormality constraint. The update can be summarized as:
\begin{equation} \label{eq:qr_retraction}
\begin{aligned}
&Y'{k} = B_k + \alpha \xi_k \\
&B_{k+1} = \text{qf}(Y'_{k}) \\ &\text{(the Q factor from QR decomposition)}
\end{aligned}
\end{equation}
Here, $\alpha$ is the step size (learning rate) and $\xi_{k}$ is the tangent vector at $B_{k}$(derived from the gradient and potentially momentum). While QR decomposition can be computationally more intensive than some other approximations for very large matrices, it provides a numerically stable and well-established method for retraction onto the Stiefel manifold.

Furthermore, to effectively apply momentum-based optimizers, such as Adam, for the optimization of $B$ on the Stiefel manifold, Stiefel-LoRA adopts a standard strategy. Momentum-related computations (e.g., updates to first and second moments in Adam) are first performed in the ambient Euclidean space using the Euclidean gradient $\nabla_{B}f$. This yields a momentum-updated Euclidean direction, let's call it $M^{'}_{k+1}$. This direction $M'_{k+1}$ is then projected onto the subsequently used in the QR-based retraction of momentum as described in Equation (\ref{eq:qr_retraction}). This approach allows the incorporation of momentum from optimizers like Adam while rigorously maintaining the manifold constraint through projection and retraction.
\section{Experiments}
\subsection{Performance Analysis}
In this section, we present a comprehensive analysis of the performance improvements achieved by applying Stiefel manifold optimization compared to traditional AdamW optimization for LoRA fine-tuning. The experiments are conducted across key NLP benchmark domains: Commonsense Reasoning, Reading Comprehension, and Mathematics. For each category, we evaluated several benchmark datasets using three model scales LLaMA-3.2-1B, LLaMA-3.2-3B, and LLaMA3-8B \cite{grattafiori2024llama} with both standard LoRA and DoRA adaptation methods. Additionally, we analyze and discuss how the geometric constraints of the Stiefel manifold influence model training efficiency and representational capacity.

\paragraph{Commonsense Reasoning}
Before examining the results, we referenced the experimental setup from \citet{hu2022lora}. As shown in Table \ref{tab:llama_commonsense}, Stiefel manifold optimization consistently demonstrates superior performance compared to AdamW across all commonsense reasoning benchmarks. The performance improvements are particularly pronounced in complex reasoning tasks that require deeper inferential capabilities, such as ARC-c \cite{allenai:arc} and HellaSwag \cite{zellers2019hellaswag}.

These findings suggest that commonsense reasoning demands efficient learning of various forms of implicit knowledge and causal relationships. LoRA adapters trained with conventional AdamW optimization appear to have either insufficiently captured these relationships or exhibited redundancy in their representations. In contrast, the \name imparts orthogonality to the LoRA adapters, particularly to the $B$ matrix, guiding each basis vector to function as an independent information channel.

This approach enables effective representation of diverse types of information and facilitates balanced reasoning. Consequently, our optimization methodology significantly enhances commonsense reasoning capabilities that leverage contextual understanding across various knowledge domains.

Furthermore, when combined with DoRA, the \name achieves optimal performance across all model scales. This indicates a complementary effect between DoRA's decomposed low-rank adaptation and the geometric constraints imposed by the \name.

\paragraph{Reading Comprehension}
\begin{table}[ht]
\centering
\vskip 0.1in
\setlength{\tabcolsep}{1.2mm}
\resizebox{0.48\textwidth}{!}{
\begin{tabular}{ccccc}
\toprule

\textbf{Model} & \textbf{Method} & \textbf{Optimizer} & \textbf{SQuAD(F1/EM)} & \textbf{QuAC(F1)}\\ \midrule
\toprule
\multirow{2}{*}{LLaMA3.2-1B} 
& \multirow{2}{*}{LoRA} & \textbf{Stiefel} & \textbf{67.9/55.7} & \textbf{50.4} \\ 
&                        & AdamW & 64.1/51.5 & 45.9 \\ \hline
\multirow{2}{*}{LLaMA3.2-3B} 
& \multirow{2}{*}{LoRA} & \textbf{Stiefel} & \textbf{80.3/72.1} & \textbf{61.8} \\ 
&                        & AdamW & 78.6/67.4 & 57.5 \\ \hline
\multirow{2}{*}{LLaMA3-8B} 
& \multirow{2}{*}{LoRA} & \textbf{Stiefel} & \textbf{88.1/79.7} & \textbf{69.7} \\ 
&                        & AdamW & 84.3/74.6 & 65.8 \\ 
\bottomrule
\end{tabular}}
\caption{Accuracy comparison of reading comprehension using each optimizers on the SQuAD and QuAC datasets.}
\label{tab:reading}
\vskip -0.1in
\end{table}

The results for Reading Comprehension tasks further validate the effectiveness of our optimization approach. Models optimized using our method demonstrate superior performance compared to those optimized with AdamW on both the SQuAD \cite{rajpurkar2016squad} and QuAC \cite{choi2018quac} datasets.

Reading Comprehension tasks demand the ability to efficiently process complex sentence structures and contextual information while performing multi-step reasoning. The QuAC dataset, which features conversational question answering requiring contextual understanding across multiple turns, suggests that models fine-tuned with our approach enhance the model's capacity to maintain consistent representations, resulting in improved contextual understanding.

For the SQuAD dataset, the performance gap between our optimization approach and AdamW increases as the model size grows, indicating that our approach scales well with larger models in extractive question answering tasks. This pattern differs from what was observed in commonsense reasoning tasks, where smaller models showed greater relative improvements.

AdamW-based models may experience interference between processing layers due to basis vectors that are not clearly differentiated. In contrast, our optimization approach uses orthogonality to enable each vector to function as an independent information processing module. This effectively separates reading comprehension sub-tasks such as key information extraction, allowing for more sophisticated learning.

\paragraph{Math}
Mathematical reasoning represents perhaps the most challenging category in our evaluation, requiring precise logical thinking and step-by-step problem solving. Our results demonstrate that Stiefel manifold optimization provides substantial improvements for these mathematically intensive tasks.
\begin{table}[ht]
\centering
\vskip 0.1in
\setlength{\tabcolsep}{1.2mm}
\resizebox{0.48\textwidth}{!}{
\begin{tabular}{ccccc}
\toprule
\textbf{Model} & \textbf{Method} & \textbf{Optimizer} & \textbf{GSM8K} & \textbf{MATH}\\ \midrule
\toprule
\multirow{2}{*}{LLaMA3.2-1B} 
& \multirow{2}{*}{LoRA} & \textbf{Stiefel} & \textbf{35.4} & \textbf{26.5} \\ 
&                        & AdamW & 20.5 & 21.4 \\ \hline
\multirow{2}{*}{LLaMA3.2-3B} 
& \multirow{2}{*}{LoRA} & \textbf{Stiefel} & \textbf{43.4} & \textbf{33.5} \\ 
&                        & AdamW & 29.1 & 27.7 \\ \hline
\multirow{2}{*}{LLaMA3-8B} 
& \multirow{2}{*}{LoRA} & \textbf{Stiefel} & \textbf{58.8} & \textbf{22.5} \\ 
&                        & AdamW & 54.7 & 19.3 \\ 
\bottomrule
\end{tabular}}
\caption{Accuracy comparison on mathematics benchmarks using each optimizers on the GSM8K and MATH datasets.}
\label{tab:math}
\vskip -0.1in
\end{table}

Performance on GSM8K \cite{cobbe2021gsm8k}, which focuses on elementary-level word problems, shows consistent improvement with Stiefel optimization across all model sizes. This suggests that our approach helps models better capture the fundamental mathematical relationships and reasoning patterns necessary for solving arithmetic word problems.

For the more challenging MATH \cite{hendrycks2021measuring} dataset, which includes advanced problems from mathematics competitions, the benefits of Stiefel optimization are even more pronounced. This is particularly true for the larger LLAMA-3-8B model, where our approach provides significant gains over AdamW optimization. This indicates that the constraints imposed by the Stiefel manifold on parameter updates are especially beneficial for preserving and enhancing the complex mathematical reasoning capabilities of larger pre-trained models.

As with other task categories, the combination of DoRA adaptation and Stiefel optimization consistently achieves the best performance across both mathematical reasoning benchmarks and all model sizes.

\subsection{Parameter Space Properties Analysis}
To better understand the success factors of our Stiefel manifold optimization approach, we conducted an analysis of parameter space properties, focusing on orthogonality and parameter efficiency.

\begin{figure}[h]
\vskip 0.2in
\begin{center}
\centerline{\includegraphics[width=1\columnwidth]{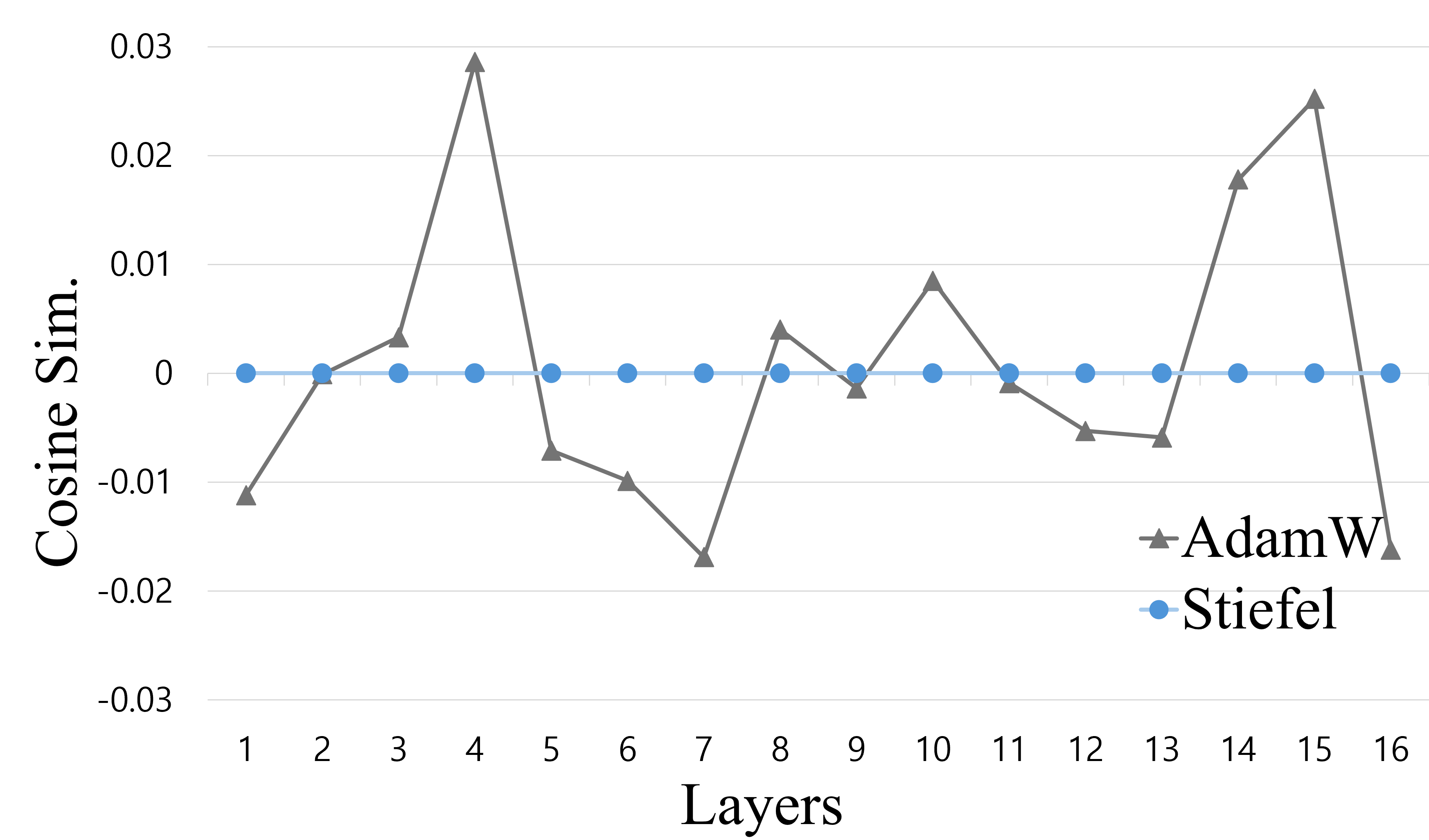}}
\caption{Comparison of mean Cosine Similarity of LoRA $B$ matrix across layers for each optimizers.}
\label{fig2}
\end{center}
\vskip -0.2in
\end{figure}

\paragraph{Cosine Similarity of Matrix $B$}
To investigate the effect of orthogonality constraints in Stiefel manifold optimization, we calculated the cosine similarity between columns of LoRA adapter matrices across different layers. Figure \ref{fig2} visualizes the cosine similarity distribution for both AdamW and Stiefel optimization after fine-tuning the LLAMA-3.2-1B model.

As expected, the Stiefel manifold optimization approach maintains perfect orthogonality, with cosine similarity values consistently maintained at zero across all layers due to the orthogonality constraints explicitly imposed during the optimization process. In contrast, AdamW exhibits varied cosine similarity values. While the mean value appears close to zero at approximately 0.003, the average standard deviation is quite large at 0.5143. This indicates that linear independence of the low-rank adaptation matrices is not guaranteed.

This analysis suggests that AdamW is an insufficient optimization method for achieving the core objective of LoRA, which is adaptation in a low-dimensional rank. The Stiefel manifold optimization approach successfully maintains orthogonal structure throughout the training process, preserving the geometric properties of the parameter space and preventing redundancy in the learned representations.

\paragraph{Effective Rank Analysis}

A critical question in LoRA fine-tuning is whether the specified rank is fully utilized during training. To investigate this, we calculated the effective rank of LoRA adapters trained with AdamW and Stiefel manifold optimization.

\begin{figure}[h]
\vskip 0.2in
\begin{center}
\centerline{\includegraphics[width=1\columnwidth]{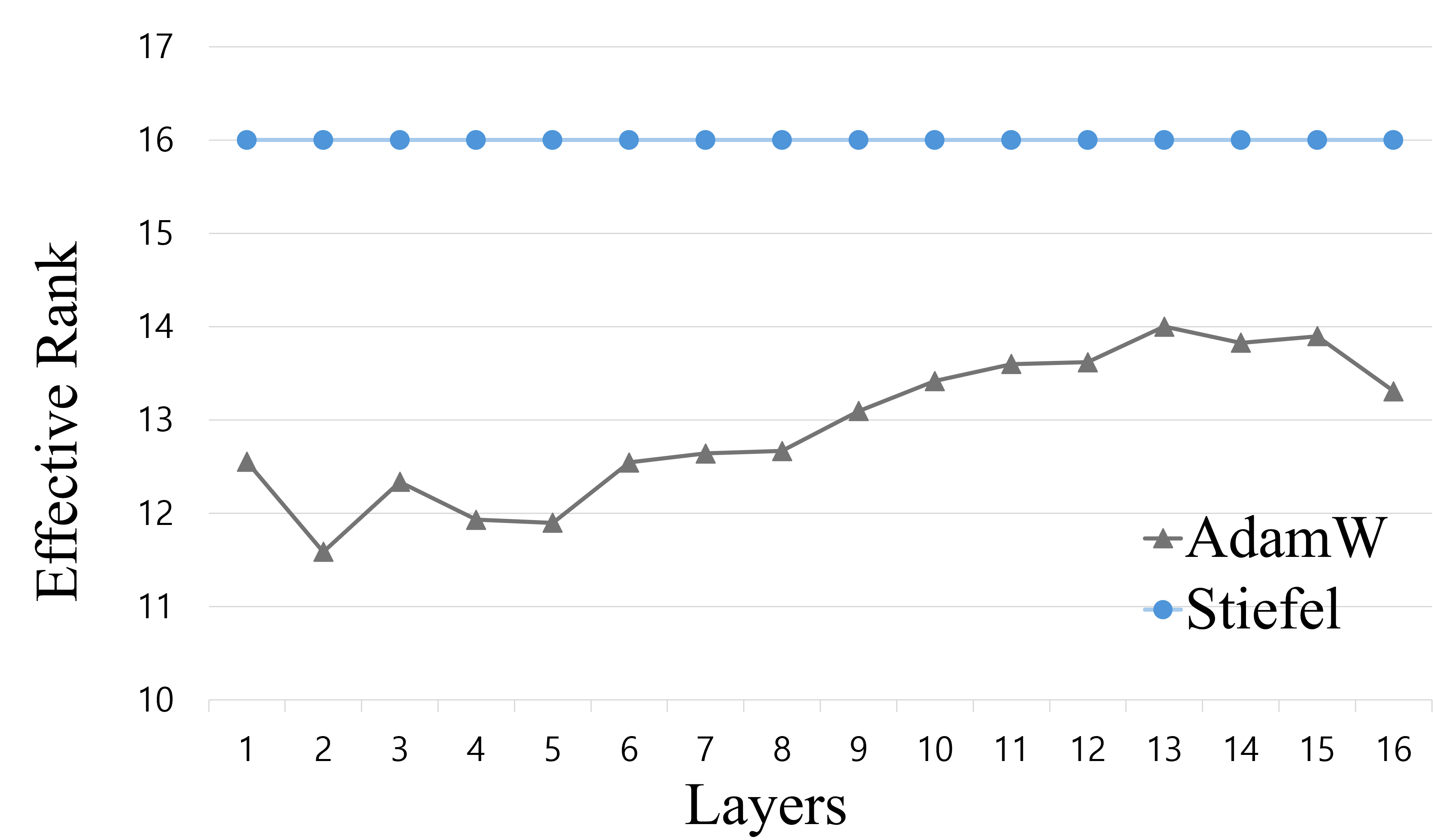}}
\caption{Layer-wise effective rank of LoRA updated matrix ($\nabla W$) for each optimizers}
\label{fig3}
\end{center}
\vskip -0.2in
\end{figure}

Figure \ref{fig3} shows the effective rank achieved by each optimization approach across layers. The results reveal notable differences. The visualization shows results for the LLaMA-3.2-1B model with a specified rank of 16. Stiefel optimization consistently utilizes all 16 dimensions fully, while AdamW effectively uses only 12 dimensions on average, failing to fully utilize the available rank space. This pattern is consistent across various rank settings, including 4, 8, 32, and 64 (See Table \ref{tab:rank_ablation}).
This inefficiency in rank utilization helps explain the performance gap between the two optimization approaches. By ensuring that all dimensions of the low-rank adaptation are effectively utilized, Stiefel optimization maximizes representational learning from limited parameters in LoRA fine-tuning.

\begin{table*}
\setlength{\tabcolsep}{1.2mm}
\centering
\begin{tabular}{ccccccccc}
\toprule
\textbf{Model} & \textbf{Method} & \textbf{Optimizer} & \textbf{BoolQ} & \textbf{PIQA} & \textbf{ARC-e} & \textbf{ARC-c} & \textbf{OBQA} & \textbf{Avg.}\\
\toprule
\midrule

\multirow{2}{*}{LLaMA3.2-1B} 
& \multirow{2}{*}{LoRA} & \textbf{Stiefel} & 68.8 & 52.1 & 46.1 & 23.9 & 59.4 & 50.1 \\
&                        & AdamW  & 64.5 & 68.3 & 63.3 & 38.5 & 50.7 & \textbf{57.1} \\ 

\bottomrule
\end{tabular}
\caption{Performance comparison of LoRA with only matrix $B$ fine-tuned using each optimizers.}
\label{tab:llama_asym}
\end{table*}

\subsection{Stiefel-LoRA with a Static Matrix $A$}
Given that Stiefel-LoRA exclusively applies its orthogonality constraint to Matrix $B$, we investigated the scenario where matrix $A$ is initialized randomly and then fixed, while only matrix $B$ is trained. This experimental setup was inspired by findings such as those in the \citet{zhu2024asymmetry}, which suggests that when the parameter budget is constrained, fine-tuning Matrix $B$ yields more significant performance improvements than fine-tuning matrix $A$. Considering the consistent performance gains demonstrated by Stiefel-LoRA over AdamW-based LoRA (as shown in previous sections), we initially anticipated similar results in this fixed-$A$ configuration. However, the results presented in Table \ref{tab:llama_asym} indicate that Stiefel-LoRA does not uniformly outperform AdamW under these specific conditions.

Our analysis suggests this outcome stems from the distinct, yet complementary, roles of matrix $A$ and matrix $B$ in learning task-specific features. While both matrices contribute, we can conceptualize matrix $A$ as primarily responsible for learning a broad set of representations relevant to the task. Matrix $B$, then, can be viewed as a specialized feature extractor that selects and refines the information learned by matrix $A$, tailoring it to the immediate input context. This nuanced view aligns with the idea that LoRA matrices act as "feature amplifiers" \cite{hu2022lora}, with matrix $B$ potentially playing a more selective role, akin to mechanisms described in works like \citet{kopiczko2023vera}, which also explore efficient adapter designs.

The critical issue arises when Matrix $A$ is a randomly initialized fixed matrix. If matrix $A$ fails to capture any meaningful or task-relevant features in its random projection, Stiefel-LoRA, applied to matrix $B$, cannot leverage its primary advantage. The orthogonality enforced by Stiefel-LoRA is designed to ensure that matrix $B$ learns to extract diverse, independent, and thereby highly informative features. However, if the input from matrix $A$ (i.e., $Ax$) is essentially noise or lacks learnable structure, the orthogonal basis of Stiefel-LoRA $B$ matrix has no meaningful signal to deconstruct and refine efficiently. Its constraint towards learning distinct features becomes less effective when there are no distinct, useful features to begin with.

In contrast, AdamW, with its greater flexibility and lack of explicit orthogonality constraints on $B$, might still identify and exploit spurious correlations or any marginal statistical regularities present in the output of the fixed random matrix $A$. This could lead to comparable or even slightly better performance in some specific instances, not because AdamW is inherently superior, but because it can adapt to the unstructured nature of the input from a fixed, random $A$.

While this behavior could be perceived as a limitation of Stiefel-LoRA, it also underscores a fundamental requirement for its optimal operation: the presence of meaningful input features from matrix $A$. As demonstrated by our comprehensive experiments where both $A$ and $B$ are trained, when Matrix A is able to learn and provide relevant information, Stiefel-LoRA constrained optimization on matrix $B$ consistently leads to superior performance by ensuring a more efficient and robust extraction and utilization of those learned features. This highlights the importance of co-adaptation of both LoRA matrices for Stiefel-LoRA to achieve its full potential.
\section{Conclusion}

This study introduced Stiefel-LoRA, a novel optimization approach for LoRA that leverages geometric constraints by optimizing its $B$ matrix on the Stiefel manifold. This explicit orthogonality enforcement aimed to enhance representational efficiency and overcome limitations of standard Euclidean optimization. Extensive experiments across diverse benchmarks and LLMs scale demonstrated that Stiefel-LoRA consistently outperformed conventional LoRA trained with AdamW.

Key contributions include the proposal and experimental validation of Stiefel-LoRA, and an analysis of internal metrics (orthogonality, effective rank) revealing the mechanism of improved representation efficiency. These findings highlight the importance of geometric constraints in PEFT design and suggest Stiefel manifold optimization as a potent method to significantly enhance PEFT performance.

\section*{Limitations}



Our study's primary limitations include the exclusive use of LLaMA series base models, thereby omitting experiments on instruction-tuned (Instruct) models prevalent in practical LLMs applications \cite{touvron2023llama2openfoundation}, and the consequent lack of qualitative analysis of generated text. Future work will aim to address these experimental gaps, further investigate the influence of the resulting independent basis vectors, and explore adaptive rank allocation methodologies, similar to approaches like \citet{zhang2023adalora}.

\bibliography{latex/custom}

\onecolumn
\appendix
\section{LoRA Evaluation at Rank $r=32$ on Seven Commonsense Reasoning Benchmarks}
\begin{table*}[h]
\setlength{\tabcolsep}{1.2mm}
\centering
\resizebox{0.99\textwidth}{!}{
\begin{tabular}{cccccccccccccc}
\toprule
\textbf{Model} & \textbf{Method} & \textbf{Optimizer} & \textbf{BoolQ} & \textbf{PIQA} & \textbf{SIQA} & \textbf{HellaSwag} & \textbf{ARC-e} & \textbf{ARC-c} & \textbf{OBQA} & \textbf{Avg.}\\
\toprule
\midrule

\multirow{4}{*}{LLaMA3.2-1B} 
& \multirow{2}{*}{LoRA} & \textbf{Stiefel} & 77.9 & 72.9 & 68.4 & 33.6 & 72.9 & 48.7 & 65.6 & \textbf{62.86} \\
&                                 & AdamW  & 65.3 & 59.3 & 58.1 & 26.5 & 61.8 & 37.1 & 50.0 & 51.16 \\ 
& \multirow{2}{*}{DoRA} & \textbf{Stiefel} & 80.5 & 73.9 & 68.2 & 33.3 & 73.7 & 40.5 & 67.8 & \textbf{62.56} \\
&                        & AdamW  & 70.4 & 67.5 & 62.1 & 29.5 & 63.8 & 38.1 & 50.0 & 54.49 \\

\midrule

\multirow{4}{*}{LLaMA3.2-3B} 
& \multirow{2}{*}{LoRA} & \textbf{Stiefel} & 86.5 & 87.1 & 83.5 & 91.5 & 87.4 & 70.7 & 82.5 & \textbf{84.17} \\
&                        & AdamW  & 83.5 & 82.7 & 79.6 & 88.1 & 85.4 & 67.7 & 78.9 & 80.84 \\
& \multirow{2}{*}{DoRA} & \textbf{Stiefel} & 88.5 & 89.4 & 85.4 & 93.7 & 89.2 & 72.6 & 84.3 & \textbf{86.16} \\
&                        & AdamW  & 85.3 & 84.7 & 81.2 & 90.4 & 87.3 & 69.5 & 80.7 & 82.73 \\

\midrule

\multirow{4}{*}{LLaMA3-8B} 
& \multirow{2}{*}{LoRA} & \textbf{Stiefel} & 90.5  & 92.3 & 88.4 & 95.7  & 92.5  & 76.2 & 87.0  & \textbf{88.94} \\
&                        & AdamW  & 88.2  & 87.9 &  84.2  & 93.6  & 90.1   & 72.4 & 83.7  & 85.73 \\
& \multirow{2}{*}{DoRA} & \textbf{Stiefel} & 92.5  & 93.8  & 89.6  & 97.7   & 93.1  & 76.5  & 88.5  & \textbf{90.24} \\
&                        & AdamW  & 89.4 & 88.3 & 85.9 & 94.4  & 91.5 & 73.6 & 84.8 & 86.84 \\

\bottomrule
\end{tabular}
}
\caption{Accuracy comparison on seven commonsense reasoning datasets with various PEFT($r=32$) method and optimizer applied.}
\end{table*}

\section{Hyperparameters of LLM Benchmarks}
\begin{table*}[ht]
\centering
\vskip 0.1in
\small
\begin{tabular}{ccccccc}
\toprule
\textbf{Hyperparameters} & \multicolumn{2}{c}{LLaMA-3.2-1B} & \multicolumn{2}{c}{LLaMA-3.2-3B} & \multicolumn{2}{c}{LLaMA3-8B} \\ \midrule
Rank r & 16 & 32 & 16 & 32 & 16 & 32  \\ 
$\alpha$ & 32 & 64 & 32 & 64 & 32 & 64 \\ 
Dropout & \multicolumn{6}{c}{0.05} \\ 
Optimizer & \multicolumn{6}{c}{AdamW}  \\ 
LR & \multicolumn{6}{c}{1e-4}  \\
LR Scheduler & \multicolumn{6}{c}{Linear} \\
Batch size & \multicolumn{6}{c}{16} \\
Warmup Steps & \multicolumn{6}{c}{100} \\
Epochs & \multicolumn{6}{c}{10} \\
Where & \multicolumn{6}{c}{Q,K,V,O,Up,Down} \\
\bottomrule
\end{tabular}
\label{tab:llama_commonsense_hyperparameters}
\vskip -0.1in
\centering
\vskip 0.1in
\small
\begin{tabular}{ccccccccc}
\toprule
\textbf{Hyperparameters} & \multicolumn{2}{c}{LLaMA-3.2-1B} & \multicolumn{2}{c}{LLaMA-3.2-3B} & \multicolumn{2}{c}{LLaMA3-8B} \\ \midrule
Rank r & 16 & 32 & 16 & 32 & 16 & 32  \\ 
$\alpha$ & 32 & 64 & 32 & 64 & 32 & 64 \\ 
Dropout & \multicolumn{6}{c}{0.05} \\ 
Optimizer & \multicolumn{6}{c}{Stiefel Manifold}  \\ 
LR & 0.3 & 0.2 & 0.3 & 0.2 & 0.3 & 0.2  \\
LR Scheduler & \multicolumn{6}{c}{Linear} \\
Batch size & \multicolumn{6}{c}{16} \\
Epochs & \multicolumn{6}{c}{10} \\
Where & \multicolumn{6}{c}{Q,K,V,O,Up,Down} \\
\bottomrule
\end{tabular}
\caption{Hyperparameter settings of LoRA(top) \& DoRA(bottom) for LLaMA-3.2-1B, LLaMA-3.2-3B and LLaMA3-8B on the commonsense reasoning tasks.}
\vskip -0.1in
\end{table*}

\begin{table*}[ht]
\centering
\vskip 0.1in
\small
\begin{tabular}{ccccccc}
\toprule
\textbf{Hyperparameters} & \multicolumn{2}{c}{LLaMA-3.2-1B} & \multicolumn{2}{c}{LLaMA-3.2-3B} & \multicolumn{2}{c}{LLaMA3-8B} \\ \midrule
Rank r & 16 & 32 & 16 & 32 & 16 & 32  \\ 
$\alpha$ & 32 & 64 & 32 & 64 & 32 & 64 \\ 
Dropout & \multicolumn{6}{c}{0.05} \\ 
Optimizer \& LR & \multicolumn{3}{c}{AdamW \& 1e-4} & \multicolumn{3}{c}{Stiefel \& 0.3}  \\ 
LR Scheduler & \multicolumn{6}{c}{Linear} \\
Batch size & \multicolumn{6}{c}{16} \\
Warmup Steps & \multicolumn{6}{c}{100 for AdamW} \\
Epochs & \multicolumn{6}{c}{10} \\
Where & \multicolumn{6}{c}{Q,K,V,O,Up,Down} \\
\bottomrule
\end{tabular}
\caption{Hyperparameter settings of LoRA for LLaMA-3.2-1B, LLaMA-3.2-3B and LLaMA3-8B on the reading comprehension tasks.}
\label{tab:llama_commonsense_hyperparameters}
\vskip -0.1in
\end{table*}

\begin{table*}[ht]
\centering
\vskip 0.1in
\small
\begin{tabular}{ccccccc}
\toprule
\textbf{Hyperparameters} & \multicolumn{2}{c}{LLaMA-3.2-1B} & \multicolumn{2}{c}{LLaMA-3.2-3B} & \multicolumn{2}{c}{LLaMA3-8B} \\ \midrule
Rank r & 16 & 32 & 16 & 32 & 16 & 32  \\ 
$\alpha$ & 32 & 64 & 32 & 64 & 32 & 64 \\ 
Dropout & \multicolumn{6}{c}{0.05} \\ 
Optimizer \& LR & \multicolumn{3}{c}{AdamW \& 1e-4} & \multicolumn{3}{c}{Stiefel \& 0.1}  \\ 
LR Scheduler & \multicolumn{6}{c}{Linear} \\
Batch size & \multicolumn{6}{c}{16} \\
Warmup Steps & \multicolumn{6}{c}{100 for AdamW} \\
Epochs & \multicolumn{6}{c}{10} \\
Where & \multicolumn{6}{c}{Q,K,V,O,Up,Down} \\
\bottomrule
\end{tabular}
\caption{Hyperparameter settings of LoRA for LLaMA-3.2-1B, LLaMA-3.2-3B and LLaMA3-8B on the mathematics tasks.}
\label{tab:llama_commonsense_hyperparameters}
\vskip -0.1in
\end{table*}

\newpage
\section{Algorithms}
\subsection{Preliminaries for Stiefel Manifold Optimization}
\label{app:preliminaries}
\newtheorem{definition}{Definition}
\begin{definition}\textbf{\textit{Low-Rank Adaptation:}}
Let $W_{\text{0}} \in \mathbb{R}^{d \times k}$ be a pre-trained weight matrix. Low-Rank Adaptation (LoRA) performs fine-tuning by freezing $W_{0}$ and adding a low-rank matrix product $BA$ to it, where $B \in \mathbb{R}^{d \times r}$ and $A \in \mathbb{R}^{r \times k}$. The rank $r$ is chosen such that $r \ll \min(d, k)$. The updated weight matrix $\textbf{W}$ is thus defined as:
\begin{equation}
    \textbf{W} = W_{\text{0}} + BA
\end{equation}
\end{definition}
In standard LoRA, only the matrices $A$ and $B$ are trainable parameters. These parameters are typically updated using standard first-order optimization algorithms, such as Adam, in Euclidean space without explicit constraints on $A$ or $B$.

\begin{definition}\textbf{\textit{Stiefel Manifold:}}
The Stiefel manifold $St(n, p)$, for integers $n \geq p$, is defined as the set of all $n \times p$ real matrices with orthonormal columns. Formally:
\begin{equation}
    St(n, p) = \{ X \in \mathbb{R}^{n \times p} \mid X^T X = I_p \}
\end{equation}
\end{definition}
where $I_p$ is the $p \times p$ identity matrix. The Stiefel manifold is a smooth differentiable manifold. In this work, we impose the constraint that the LoRA matrix $B \in \mathbb{R}^{d \times r}$ lies on the Stiefel manifold.

\begin{definition}\textbf{\textit{Concepts in Riemannian Optimization:}}
Riemannian optimization refers to the process of optimizing functions defined on Riemannian manifolds. Key concepts include:
\begin{itemize}
    \item[\textbf{(a)}] \textbf{Tangent Space:} For a point $X$ on a manifold $M$, the tangent space $T_X M$ is a vector space consisting of all possible directions (tangent vectors) one can move from $X$ while staying on $M$.

    \item[\textbf{(b)}] \textbf{Riemannian Gradient:} Given a differentiable function $f: M \to \mathbb{R}$ on a Riemannian manifold $M$, the Riemannian gradient $\text{grad} f(X)$ at a point $X \in M$ is an element of the tangent space $T_X M$. If $M$ is embedded in a Euclidean space, the Euclidean gradient $\nabla f(X)$ (gradient in the ambient space) generally does not belong to $T_X M$. The Riemannian gradient is then obtained by projecting the Euclidean gradient onto the tangent space:
    \begin{equation}
        \text{grad} f(X) = \text{proj}_{T_X M}(\nabla f(X))
    \end{equation}
    where $\text{proj}_{T_X M}(\cdot)$ (or $\pi_{T_X M}(\cdot)$ as in the provided text) denotes the orthogonal projection onto $T_X M$.

    \item[\textbf{(c)}] \textbf{Retraction:} A retraction $R_X: T_X M \to M$ is a mapping from the tangent space $T_X M$ to the manifold $M$. For a tangent vector $\eta_X \in T_X M$, $R_X(\eta_X)$ provides a new point on the manifold $M$ by moving from $X$ in the direction $\eta_X$. This serves as an update step in an optimization algorithm on $M$. A retraction must satisfy $R_X(0_X) = X$ (where $0_X$ is the zero vector in $T_X M$) and the derivative of $R_X$ at $0_X$, $DR_X(0_X)$, must be the identity map on $T_X M$.

    \item[\textbf{(d)}] \textbf{Vector Transport:} Vector transport $\mathcal{T}_{\eta_X}(\xi_X)$ is a process that moves a tangent vector $\xi_X \in T_X M$ along a direction $\eta_X \in T_X M$ to the tangent space $T_{R_X(\eta_X)}M$ at the point $R_X(\eta_X)$ on the manifold. This is essential for adapting information from previous optimization steps, such as momentum, to the current step's tangent space.
\end{itemize}
\end{definition}
It is noted that standard geometric operations such as the exponential map, parallel transport, and SVD-based projections can be computationally expensive, posing challenges for direct application to large-scale deep learning models.
\newpage
\subsection{Update Parameters of Stiefel Manifold Optimization}

\newcommand{\stiefel}[2]{\mathrm{St}(#1, #2)}
\newcommand{\tr}{\mathrm{Tr}}
\newcommand{\sym}[1]{\mathrm{sym}(#1)} 
\newcommand{\qf}[1]{\mathrm{qf}(#1)}   


\begin{algorithm*}
\caption{Stiefel-LoRA Parameter Update using QR-Retraction}
\label{alg:stiefel_lora_qr_adam}
\begin{algorithmic}[1]
\State \textbf{Input:}
\State \quad Pre-trained weights $W_0$
\State \quad Initial LoRA matrices $A_0 \in \mathbb{R}^{r \times k}$, $B_0 \in \stiefel{d}{r}$ (i.e., $B_0^{\top} B_0 = I_r$)
\State \quad Learning rate for A: $\eta_A > 0$
\State \quad Step size (learning rate) for B on manifold: $\alpha_B > 0$
\State \quad Adam hyperparameters for A: $\beta_{1A}, \beta_{2A} \in [0, 1)$, $\epsilon_A > 0$
\State \quad Adam hyperparameters for B (Euclidean part): $\beta_{1B}, \beta_{2B} \in [0, 1)$, $\epsilon_B > 0$
\State \quad Number of training iterations: $T_{max}$

\State \textbf{Initialize:}
\State \quad $A \leftarrow A_0$, $B \leftarrow B_0$
\State \quad Adam first moments for A: $m_A \leftarrow 0$, for B: $m_B \leftarrow 0$
\State \quad Adam second moments for A: $v_A \leftarrow 0$, for B: $v_B \leftarrow 0$
\State \quad Iteration counter $t \leftarrow 0$

\While{$t < T_{max}$}
    \State $t \leftarrow t + 1$
    \State Compute loss $L_t = f(W_0 + B A)$
    \State Compute Euclidean gradients: $g_A \leftarrow \nabla_A L_t$, $g_B \leftarrow \nabla_B L_t$

    \State \Comment{\textit{Update matrix A (Standard Adam)}}
    \State $m_A \leftarrow \beta_{1A} m_A + (1-\beta_{1A}) g_A$
    \State $v_A \leftarrow \beta_{2A} v_A + (1-\beta_{2A}) g_A^2$ 
    \State $\hat{m}_A \leftarrow m_A / (1-\beta_{1A}^t)$
    \State $\hat{v}_A \leftarrow v_A / (1-\beta_{2A}^t)$
    \State $A \leftarrow A - \eta_A \hat{m}_A / (\sqrt{\hat{v}_A} + \epsilon_A)$ 

    \State \Comment{\textit{Update matrix B (Stiefel Manifold Optimization with QR-Retraction)}}
    \State \Comment{1. Compute Euclidean Adam preconditioned gradient direction $M'_B$}
    \State $m_B \leftarrow \beta_{1B} m_B + (1-\beta_{1B}) g_B$
    \State $v_B \leftarrow \beta_{2B} v_B + (1-\beta_{2B}) g_B^2$ 
    \State $\hat{m}_B \leftarrow m_B / (1-\beta_{1B}^t)$
    \State $\hat{v}_B \leftarrow v_B / (1-\beta_{2B}^t)$
    \State $M'_B \leftarrow \hat{m}_B / (\sqrt{\hat{v}_B} + \epsilon_B)$ \Comment{Euclidean preconditioned gradient}

    \State \Comment{2. Project $M'_B$ onto the tangent space at $B$ to get tangent vector $\xi$}
    \State $\sym{X} \triangleq (X + X^T)/2$
    \State $\xi \leftarrow M'_B - B \cdot \sym{B^T M'_B}$ \Comment{Project $M'_B$ to $T_B St(d,r)$}
    
    \State \Comment{3. Perform retraction step using QR decomposition}
    \State $Y' \leftarrow B - \alpha_B \xi$ \Comment{Step in the tangent direction (descent)}
    \State $B \leftarrow \qf{Y'}$ \Comment{Update B with the Q factor of QR decomposition of $Y'$}
\EndWhile

\State \textbf{Output:} Optimized LoRA matrix $A, B$
\end{algorithmic}
\end{algorithm*}


\newpage
\subsection{Effective Rank Calculation Algorithm}



\begin{algorithm*}
\caption{Effective Rank Calculation based on Shannon Entropy}
\label{alg:effective_rank}
\begin{algorithmic}[1]
\State \textbf{Input:} Matrix $M \in \mathbb{R}^{m \times n}$
\State \quad \quad \quad (Optional) Small constant $\epsilon > 0$ for numerical stability (e.g., $10^{-9}$)
\State \textbf{Output:} Effective Rank $R_{eff}(M)$

\State \Comment{Step 1: Perform Singular Value Decomposition (SVD)}
\State Compute SVD of $M$: $M = U \Sigma V^T$
\State Let $S = \{\sigma_1, \sigma_2, \dots, \sigma_p\}$ be the set of singular values from $\Sigma$, where $p = \min(m, n)$.
\State Ensure singular values are non-negative: $\sigma_i \ge 0$.

\State \Comment{Step 2: Filter and normalize positive singular values}
\State Let $S^{+} = \{\sigma_i \in S \mid \sigma_i > \epsilon \}$ be the set of positive singular values significantly greater than zero.
\State Let $k = |S^{+}|$ be the number of such positive singular values.

\If{$k = 0$}
    \State $R_{eff}(M) \leftarrow 0$ \Comment{Matrix is effectively a zero matrix or rank is negligible}
\Else
    \State Calculate the sum of positive singular values: $\Sigma_{\sigma} = \sum_{\sigma_j \in S^{+}} \sigma_j$.
    \If{$\Sigma_{\sigma} < \epsilon$} \Comment{Sum is too small, treat as zero rank}
        \State $R_{eff}(M) \leftarrow 0$
    \Else
        \State Normalize the positive singular values to form a probability distribution $P = (p_1, p_2, \dots, p_k)$:
        \State $p_j \leftarrow \frac{\sigma_j}{\Sigma_{\sigma}}$ for each $\sigma_j \in S^{+}$.
        
        \State \Comment{Step 3: Calculate Shannon Entropy}
        \State $H(P) \leftarrow - \sum_{j=1}^{k} p_j \ln(p_j)$
        \State \Comment{Convention: $0 \ln 0 = 0$.}

        \State \Comment{Step 4: Calculate Effective Rank}
        \State $R_{eff}(M) \leftarrow \exp(H(P))$
    \EndIf
\EndIf

\State \Return $R_{eff}(M)$
\end{algorithmic}
\end{algorithm*}

\section{Average Effective Rank for Each Rank}
\begin{table*}[h]
\centering
\caption{Hyperparameter settings of LoRA \& DoRA for LLaMA-3.2-1B, LLaMA-3.2-3B and LLaMA3-8B on the commonsense reasoning tasks.}
\vskip 0.1in
\small
\begin{tabular}{ccccccccccc}
\toprule
\multirow{2}{*}{\textbf{Model}} & \multicolumn{2}{|c|}{\textbf{4}} & \multicolumn{2}{c|}{\textbf{8}} & \multicolumn{2}{c|}{\textbf{16}} & \multicolumn{2}{c|}{\textbf{32}} & \multicolumn{2}{c|}{\textbf{64}} \\
& Stiefel & AdamW & Stiefel & AdamW & Stiefel & AdamW & Stiefel & AdamW & Stiefel & AdamW \\ 
\midrule
LLaMA-3.2-1B & 4 & 2.8 & 7.9 & 5.4 & 16 & 12.1 & 31.8 & 23.8 & 64 & 49.7 \\ 
LLaMA-3.2-3B & 4 & 3.4 & 8 & 6.2 & 16 & 13.5 & 32 & 28.6 & 64 & 55.1 \\ 
\bottomrule
\end{tabular}
\label{tab:rank_ablation}
\vskip -0.1in
\end{table*}

\end{document}